\documentclass[pdflatex,sn-mathphys-num]{sn-jnl}

\usepackage{graphicx}
\usepackage{multirow}
\usepackage{amsmath,amssymb,amsfonts}
\usepackage{amsthm}
\usepackage{mathrsfs}
\usepackage[title]{appendix}
\usepackage{xcolor}
\usepackage{textcomp}
\usepackage{manyfoot}
\usepackage{booktabs}
\usepackage{algorithm}
\usepackage{algorithmicx}
\usepackage{algpseudocode}
\usepackage{listings}
\usepackage{longtable}
\usepackage{array}
\usepackage{calc}
\usepackage{ragged2e}
\usepackage{enumitem}
\usepackage{url}
\usepackage{hyperref}
\usepackage{tabularx}
\usepackage{pdflscape}

\begin{document}

\title[Urdu Automatic Speech Recognition Survey]{From Statistical Methods to Pre Trained Models; A Survey on Automatic Speech Recognition for Resource Scarce Urdu Language}

\author*[1]{\fnm{Muhammad} \sur{Sharif}}\email{msharifwali@hotmail.com}
\author*[2]{\fnm{Sardar Shan Ali} \sur{Naqvi}}\email{d\_shanali@hust.edu.cn}
\author[3]{\fnm{Zeeshan} \sur{Abbas}}\email{zeeshanabbas23@email.szu.edu.cn}
\author[1]{\fnm{Wu} \sur{Ying}}\email{wuyingivy@szu.edu.cn}

\affil*[1]{\orgname{College of International Studies, Shenzhen University, Nanhai Avenue, Nanshan District, Shenzhen, 518060, China}}
\affil*[2]{\orgname{College of Artificial Intelligence and Automation, Huazhong University of Science and Technology, 1037 Luoyu Road, Wuhan 430074, Hubei, China}}
\affil[3]{\orgname{School of Electronics and Information Engineering, Shenzhen University, Nanhai Avenue, Nanshan District, Shenzhen, 518060, China}}

\abstract{Automatic speech recognition (ASR) has advanced rapidly and has improved human--computer interaction. However, low-resource languages such as Urdu have benefited less from these developments because of limited speech data, text resources, and research tools. This paper surveys the development of Urdu ASR from statistical methods to deep neural networks, end-to-end systems, and pre-trained multilingual models. It reviews the available Urdu speech datasets, feature extraction methods, recognition tools, model architectures, and evaluation measures. It also compares monolingual and multilingual approaches and discusses the main limitations of existing studies. The review shows that small and task-specific datasets have restricted the use of data-intensive models, while transfer learning and multilingual pre-training provide useful directions for future research. Finally, the paper identifies open problems related to data availability, dialect variation, code-switching, and consistent performance evaluation.}

\keywords{Automatic speech recognition, low-resource language, pre-trained models, Urdu ASR, Urdu speech datasets}

\maketitle

\section{INTRODUCTION}\label{sec:introduction}

Automatic speech recognition (ASR) converts an input speech signal into written text. The speech signal is first digitized at a suitable sampling rate, after which relevant acoustic features are extracted. Common feature representations include Mel-frequency cepstral coefficients (MFCCs), filter-bank features (FBANKs), and the short-time Fourier transform (STFT) \cite{sharma, salau, kanabur, ghahramani, xu2020discriminative}. The Fourier transform represents a speech signal in the frequency domain and supports the analysis of its spectral properties \cite{osgood}. MFCCs are obtained by applying a discrete cosine transform (DCT) to the logarithmic energies of the Mel-scaled filter banks \cite{alim}. The extracted features are then processed by the recognition model to identify phonetic units, words, or complete text sequences \cite{nakatani}.

Automatic speech recognition (ASR) systems have been widely researched and developed for various languages and domains. However, the development of ASR systems for resource-scarce languages \cite{diwan}, such as Urdu, has been limited due to the availability of limited annotated speech data and computational resources \cite{farooq2019dnn, zia2018lexicon, reitmaier, naeem}. Statistical techniques like HMM--GMM are widely adopted for small-scale datasets across languages with varied accuracies \cite{amoolya, ashraf, zhang2017exploitation}. Prior work on acoustic feature selection \cite{bhardwaj2023enhanced,mondol2026explainability} and foundational speech recognition architectures \cite{bhardwaj2020development} highlights the importance of robust pattern extraction, particularly when operating under unconstrained or acoustic-degraded conditions.

Recent advances in deep neural networks (DNNs) have increased research on ASR for low-resource languages such as Urdu \cite{farooq2019dnn, zia2018lexicon, chandio}. DNNs can model complex speech patterns effectively when sufficient training data are available \cite{maas}. However, their performance may decrease when they are trained on small datasets \cite{sailor, amoolya}. This issue is important for Urdu because large and well-annotated speech datasets remain limited. Consequently, some studies have focused on restricted tasks, such as Urdu digit recognition \cite{chandio}. Other studies have applied DNNs to the acoustic and language modeling components of Urdu ASR systems \cite{farooq2019dnn, khan}. A hybrid DNN-HMM system has been used for continuous Urdu speech recognition, while other studies have combined neural acoustic models with n-gram language models for smaller recognition tasks \cite{kannan, humayun}. These findings suggest that DNNs can support Urdu ASR, particularly when they are combined with suitable language models \cite{xu2020discriminative, billa, fantaye, khan}. However, further work is required to improve continuous speech recognition and model generalization across different speakers, accents, and speaking styles.

The remainder of this paper is organized as follows. Section~\ref{sec:background} presents the background and literature review. Section~\ref{sec:methodology} describes the review methodology. Section~\ref{sec:urdu-asr} discusses Urdu datasets, tools, and monolingual and multilingual ASR approaches. Section~\ref{sec:comparison} compares the reviewed methods. Sections~\ref{sec:discussion} and~\ref{sec:open-problems} present the discussion, limitations, and open research problems. The paper is concluded in Section~\ref{sec:conclusion}.

\subsection{Motivation}\label{subsec:motivation}

This survey examines the main research trends in automatic speech recognition for Urdu. Although Urdu is widely spoken in South Asia, fewer speech datasets and ASR systems are available for Urdu than for major languages such as English and Mandarin \cite{metcalf, abbas, rahman, russell, vanesch}. Because of the limited availability of annotated speech data, many Urdu ASR studies have used conventional statistical methods or small, task-specific datasets \cite{chandio, qasim2016district, raza2009phonetically}. More recent studies have explored DNNs, end-to-end architectures, self-supervised learning, and pre-trained multilingual models \cite{farooq2019dnn, humayun, javed, bhogale2023mining}. However, the use of these methods for Urdu remains limited compared with their use for better-resourced languages. Considering the facts, we will be exploring and setting out to answer the following questions in this study:

\begin{enumerate}[label=\roman*.]
    \item Which methods have been used for Urdu ASR with small- and large-scale datasets?
    \item To what extent has Urdu been studied in ASR research?
    \item Which Urdu speech datasets are available, and how have they been used?
    \item Which practices can improve the performance of Urdu ASR systems?
    \item Which open problems should be addressed in future Urdu ASR research?
\end{enumerate}


\subsection{Significance of the study}\label{subsec:significance}

Considering the already stated background in multiple aspects and keeping the objectives in mind, it is imperative to disclose the facts and ongoing trends in ASR research for the sake of setting future research needs and inquiring about the spaces left. There have been limited attempts to review Urdu ASR systems collectively. To the best of our knowledge, this work is one of the few focused surveys of automatic speech recognition for Urdu, a widely spoken language across South Asia. It summarizes the available speech datasets, adopted technologies, monolingual and multilingual approaches, evaluation methods, and major research challenges in Urdu language automatic speech recognition.


\section{BACKGROUND OF THE STUDY AND LITERATURE REVIEW}\label{sec:background}


\subsection{Definition and overview of automatic speech recognition (ASR)}\label{subsec:definition}

Automatic speech recognition is the ability of a computer system to convert spoken language into text. ASR has been studied for several decades and has progressed considerably because of advances in machine learning and deep neural networks \cite{baker, alam}. However, the lack of suitable speech corpora remains a major challenge for many low-resource languages \cite{besacier}.

Urdu is spoken by more than 100 million people, mainly in South Asia \cite{daud, russell, rahman, oesterheld}. Despite its large number of speakers, Urdu has relatively few large and publicly available speech corpora. Many existing datasets contain isolated words, digits, read speech, or speech from a limited number of speakers \cite{qasim2016travel, ali2015pitch, chandio}. This limited availability of diverse and annotated data makes it difficult to develop Urdu ASR systems that perform well across different speakers and recording conditions. Complementary to end-to-end pre-trained neural networks, standard speech signal processing benchmarks \cite{bhardwaj2021effect,alkhayyat2026feature} demonstrate that acoustic feature stability remains crucial for mitigating noise and domain shifts in low-resource speech environments.


\subsection{Impact of Resource Scarcity and Approaches to Improve ASR Performance}\label{subsec:resource-scarcity}

ASR researchers have used several approaches, including Gaussian mixture model-hidden Markov model (GMM-HMM) systems, deep neural networks (DNNs), and end-to-end models. DNNs often perform better than statistical models when sufficient training data are available \cite{farooq2019dnn, naeem, reitmaier}. However, their performance depends strongly on the size, quality, and diversity of the training dataset.

This limitation is important for Urdu ASR because many available datasets are small, task-specific, or not publicly available. Larger datasets generally support better recognition performance in both traditional and end-to-end systems \cite{stoian}. In low-resource settings, researchers use data augmentation, transfer learning, and pre-training to reduce the effect of limited labeled data \cite{xu2020lrspeech, besacier}. The effectiveness of these methods still depends on the quality and diversity of the original speech data.

\textbf{Deep Neural Networks (DNNs):} DNN-based approaches are extensively adopted in ASR systems, where hierarchical representations of speech features are utilized in modeling the acoustic properties. Deep feedforward networks and convolutional neural networks (CNNs) in DNN architecture capture complex patterns and variations in the acoustic data, which helps getting boosted recognition accuracy given a high volume training datasets \cite{amoolya}.

\textbf{Recurrent Neural Networks (RNNs):} RNNs and its variants like long short-term memory (LSTM) and gated recurrent unit (GRU) are adopted for sequential data processing, appropriately modelling the temporal dependencies in speech \cite{shewalkar}. Literature \cite{oruh} states that the RNN-based techniques are mainly helpful to capture contextual information and long-term speech sequence dependencies, hence improve the ASR performance.

\textbf{Attention Mechanisms:} Lengthy utterances and relevant contexts from speech patterns are specifically managed by the Attention mechanism. It pays attention on input segments diverting the focus of ASR system during the decoding process \cite{chiu}. Sequence to sequence and other variants of attention mechanism has served different ASR systems \cite{mandava}.

\textbf{Data Augmentation:} Additional training data is generated by data augmentation technique by perturbation of speech, injecting noise, shifting speech pitch, and transforming the existing limited-scale speech data \cite{xu2020lrspeech}. This is a unique way of faking the dataset helpful in enhancing the robustness and generalizability of the ASR system.

\textbf{Transfer Learning and Pre-training:} Beside resource availability, another approach to improve ASR performance is adopting technology that uses features of another language, having larger datasets \cite{kermanshahi, khare, joshi}. Transfer learning and its variants like fine-tuning pre-trained models produce sophisticated results in ASR tasks, alleviating the issue of data scarcity. In this mechanism, Pre-training large neural networks on large-scale speech datasets from rich resource languages are carried and then adapt them to the ASR tasks of low resource languages \cite{radford}. A study by \cite{mohiuddin} shows Urdu ASR can be benefitted this way. Such strategies can be adopted for Urdu ASR in a larger scale to leverage pre-trained models and enhance performance. Transfer learning techniques and pre-trained models are effective in several low-resource language settings, including Urdu. By leveraging pre-trained models ASR performance have achieved human-level accuracy \cite{radford, bhogale2023mining, javed}. The broader value of adapting a pre-trained model to a new data representation has also been demonstrated in other classification tasks with limited domain-specific data \cite{naqvi2024ddosVi,sharif2025balti}.

In addition to the choice of sophisticated algorithms, models, and efficient tools, the performance of an ASR system is also improved by the unsupervised learning paradigm \cite{chen2023focal}. The literature outlines that techniques help overcome the challenges posed by limited training data and improve the robustness and accuracy of the ASR system \cite{bai}. Similar benefits have been reported in other low-label learning tasks, where multilevel autoencoders, reconstructive models, and self-supervised prediction were used to learn useful representations from limited or unlabeled data~\cite{ali2019learning, naqvi2024ddos, naqvi2026online}.

The reviewed work shows that the development of Urdu ASR is closely linked to the available speech resources. Statistical models remain useful for small and task-specific datasets. DNN and end-to-end systems are more suitable when larger labeled datasets are available. Pre-trained multilingual models provide another direction because they transfer speech knowledge from larger datasets to Urdu. Therefore, this survey studies Urdu ASR through five connected elements: dataset size, speech type, model family, training strategy, and evaluation measure.


\subsection{Speech Recognition Frameworks}\label{subsec:frameworks}

\begin{figure}[htbp]
\centering
\includegraphics[width=0.95\textwidth]{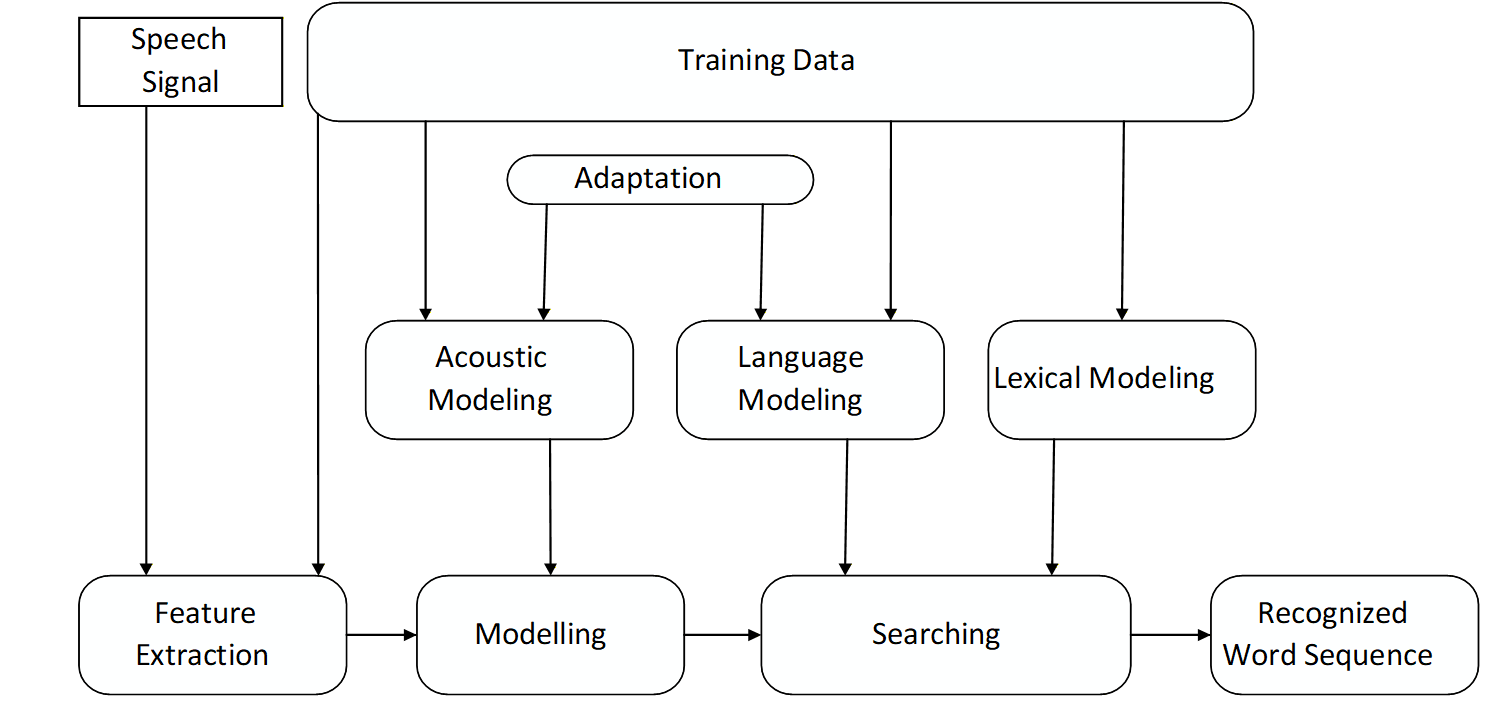}
\caption{Main modules and information flow of a conventional ASR system}\label{fig:fig1}
\end{figure}

The effectiveness of traditional generative methods and tools in building state-of-the-art automatic speech recognition (ASR) systems depends on several factors such as the size and quality of the training data, the complexity of the acoustic and language model, and the available relevant computational. On the other hand, deep learning techniques have attracted much attention because of their outstanding performance. Major deep learning approaches like HMM-DNN and end-to-end E2E models, are discriminative. Unlike the generative model, the discriminative models successfully actuated CNN and RNN, which proved to be more effective in dealing with the speech signal property of temporal invariance. The main modules of a conventional ASR system are presented in Fig.~\ref{fig:fig1}. The input speech signal first passes through feature extraction, where the waveform is converted into acoustic feature vectors. During training, these features and their corresponding transcriptions are used to build the acoustic, lexical, and language models. The acoustic model relates speech features to basic sound units. The lexical model maps these sound units to words, while the language model estimates the probability of possible word sequences. During recognition, the search module combines the outputs of these models and selects the most likely word sequence. In Urdu ASR, limited transcribed speech affects acoustic model training, while limited pronunciation and text resources also affect lexical and language modeling.

Some of the most popular tools effectively adopted in research and development of ASR systems include CMU Sphinx \cite{shaukat, qasim2016travel}, Kaldi \cite{khan, farooq2020code}, SiriKit \cite{chandio}, SAPI by Microsoft and Google Speech Recognition API, and ESPNet \cite{watanabe2018espnet}; meanwhile, PyTorch, Chainer, Tensor Flow, and Caffe has been essential ingredients to make the deep learning paradigm. PyTorch and TensorFlow are deep learning frameworks gained popularity in recent years due to their ease of use and flexibility. Caffe is another deep learning framework that is known for its efficiency in training large-scale neural networks. ESPnet is a comprehensive speech processing toolkit that places its primary emphasis on End-to-End Automatic Speech Recognition (ASR) and End-to-End Text-to-Speech (TTS) applications. It utilizes Chainer and PyTorch as its deep learning backends while adhering to the recipe structure of KALDI bash scripts. Thanks to ESPnet\textquotesingle s handling of the deep learning backend and the straightforward nature of its End-to-End approach, developing a speech system for a new language is comparatively straightforward. Table~\ref{tab:tools} enlists some of the frameworks and Tools that have been adopted in Urdu ASR by different researchers.


\begin{table}[htbp]
\centering
\caption{Software tools utilized for Urdu ASR}
\label{tab:tools}
\scriptsize
\setlength{\tabcolsep}{3pt}
\renewcommand{\arraystretch}{1.5}
\begin{tabular}{@{}p{0.6cm} p{3cm} p{4cm} p{1.8cm} p{0.7cm}@{}}
\toprule
\textbf{S. No} & \textbf{Tool(s)} & \textbf{Technology} & \textbf{Accuracy WER} & \textbf{Ref.} \\
\midrule
1 & Scikit, Keras & SVM, CNN & 0.97 & \cite{chandio} \\
2 & libSVM & SVM, RF, LDA & 73\% & \cite{ali2015pitch} \\
3 & Sphinx, CLTK & HMM & 74\% & \cite{shaukat} \\
4 & Sphinx & SVM, GMM & 92.4\% & \cite{qasim2016travel} \\
5 & Kaldi, SRILM & SGMM, MMI & 24.2 & \cite{raza2018telephonic} \\
6 & Sphinx4 & HMM & 10.66 & \cite{ashraf} \\
7 & Kaldi, KenLM & SGMMI & 16.7 & \cite{naeem} \\
8 & Kaldi, SRILM & TDNN, GMM-HMM & 18.59 & \cite{khan} \\
9 & Kaldi, SRILM & DNN-HMM,RNNLM & 24.5 & \cite{farooq2020code} \\
10 & Kaldi & TDNN,DNN, BLSTM & 13.5 & \cite{farooq2019dnn} \\
11 & - & SSL, LLE & 34.65 & \cite{humayun} \\
12 & S3prl & Wav2Vec2, SSL & 12.6 & \cite{javed} \\
13 & MATLAB & FFT, Simulink & Up to 100\% & \cite{hasnain} \\
14 & XLSR- Wav2vec2.0 & Transformers, Fine-tuning & 0.49 & \cite{mohiuddin} \\
15 & Whisper Large-V2 & Transformer, Seq-to-Seq & 22.6 & \cite{radford} \\
\botrule
\end{tabular}
\end{table}

The current application of ASR that yields high-accuracy results consist of adopting semi-supervised learning algorithms, mainly with the Transformer model. This paradigm is seen in major research works like Open AI's speech recognition model Whisper \cite{radford}, conversational AI Model LaMDA by Google \cite{thoppilan}, and Open Pre-Trained Transformer Language Model OPT by Meta AI \cite{zhang2022opt}. Utilizing the features and pre-trained models of rich resource languages, normally multilingual ASR models, fine-tuning is the process of giving the advantage of getting better results for low-resource languages \cite{bai}.


\subsection{Performance Evaluation}\label{subsec:evaluation}

The performance of an ASR system is commonly evaluated using the word error rate (WER). WER measures the difference between a reference transcription and the transcription produced by the ASR system. It considers the number of substituted words \(S\), deleted words \(D\), and inserted words \(I\), relative to the total number of words \(N\) in the reference transcription. It is calculated as:

\begin{equation}
\mathrm{WER}(\%) = \frac{S+D+I}{N}\times 100
\label{eq:wer}
\end{equation}

A lower WER indicates better recognition performance. However, WER values should be compared carefully when studies use different datasets, speech types, or evaluation settings. Phoneme error rate (PER) and character error rate (CER) follow the same edit-distance principle but use phonemes and characters as their units of evaluation. PER supports the analysis of phoneme-level recognition, whereas CER is useful for languages or systems that operate at the character level. Processing time, memory use, and real-time factors may also be reported when the computational efficiency of an ASR system is evaluated.


\section{METHODOLOGY}\label{sec:methodology}

This study follows a structured review process to examine the development of automatic speech recognition for Urdu. The review covers three main types of work. The first type includes studies that develop and evaluate Urdu ASR systems. The second includes studies that introduce speech datasets suitable for Urdu ASR. The third includes multilingual or pre-trained ASR studies that report Urdu-specific results.

The main search phrase was ``Urdu speech recognition.'' Related search combinations included ``Urdu automatic speech recognition,'' ``Urdu speech corpus,'' ``Urdu ASR dataset,'' ``Urdu wav2vec,'' ``Urdu XLSR,'' and ``Urdu Whisper.'' Journal articles, conference papers, and other studies containing sufficient technical information were considered. The reference lists of relevant papers were also examined to identify earlier Urdu ASR studies.

A study was included when it reported an Urdu speech dataset, an Urdu ASR experiment, or Urdu-specific results from a multilingual system. Dataset studies were also included when the presented Urdu speech data could support ASR research, even if ASR was not the main task of the original work. Duplicate papers, nontechnical articles, and studies that only mentioned Urdu without reporting a dataset, method, or result were excluded.

The selected papers were examined using a common set of technical fields. These fields included the dataset name, dataset size, speech type, number of speakers, public availability, feature extraction method, model architecture, software tool, evaluation measure, and reported result. This information was used to prepare the summary and comparison tables in this survey.

\begin{figure}[htbp]
\centering
\includegraphics[width=0.98\textwidth]{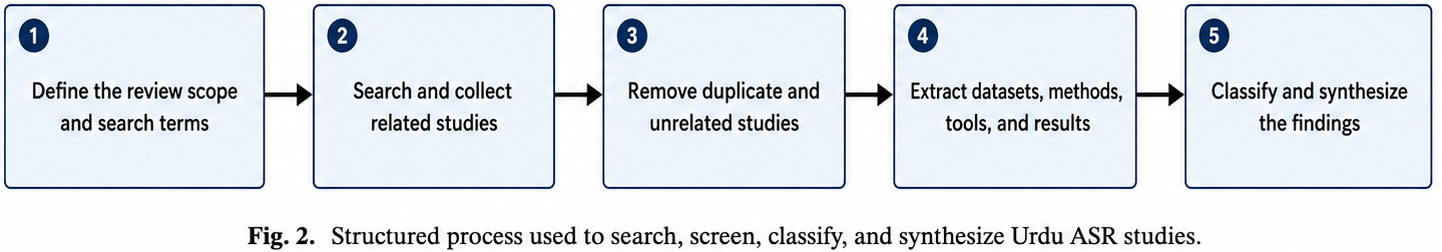}
\caption{Structured process used to search, screen, classify, and synthesize Urdu ASR studies.}\label{fig:fig2}
\end{figure}

\begin{figure}[htbp]
\centering
\includegraphics[width=1\textwidth]{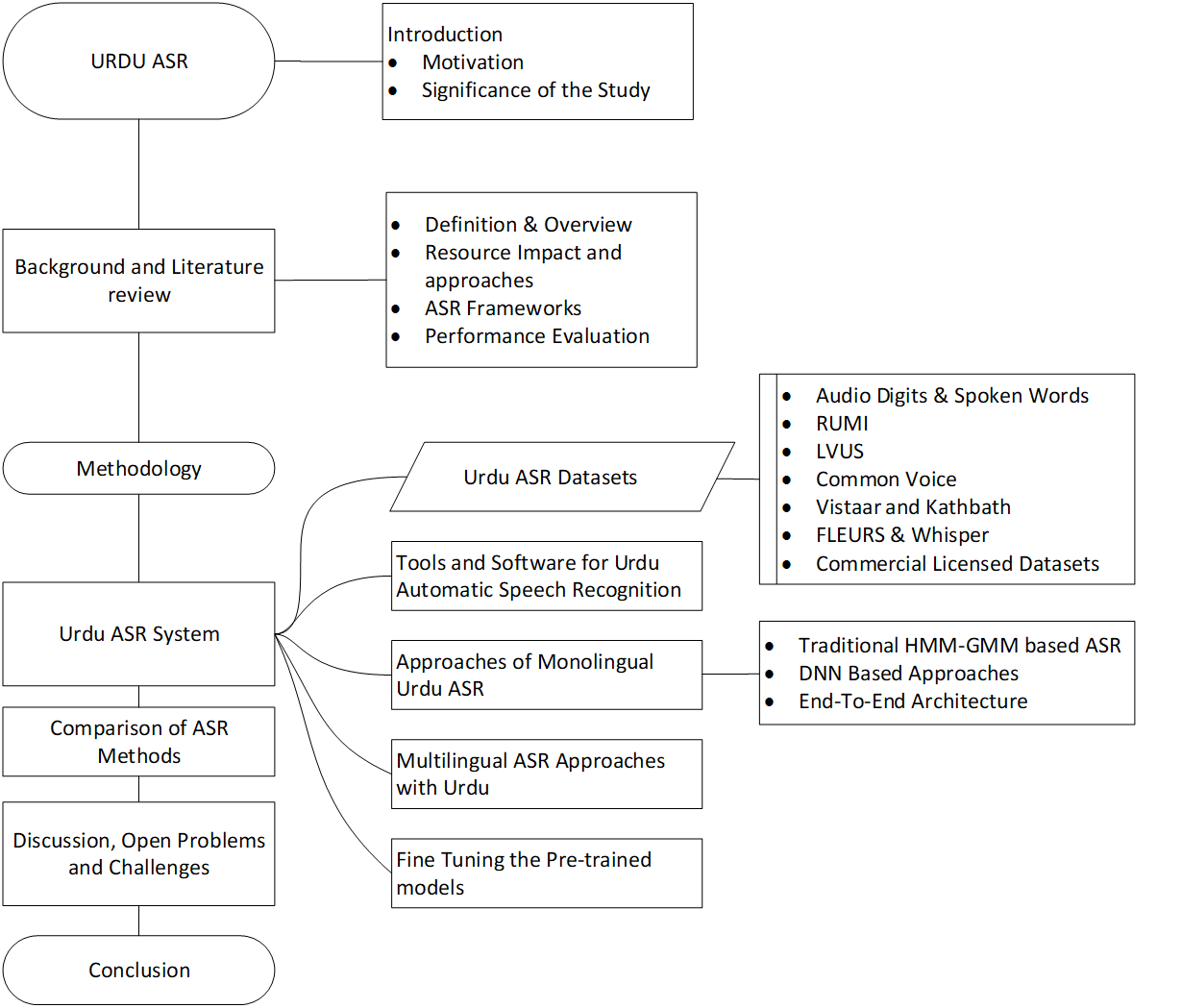}
\caption{Flow of the Study}\label{fig:fig3}
\end{figure}

The studies were then grouped into statistical, DNN-based, end-to-end, and pre-trained or multilingual approaches. The analysis considered the relation between the available data and the selected model. Studies using different datasets or evaluation measures were not treated as direct comparisons. A similar dataset-aware and method-aware comparison has been followed in recent systematic reviews of data-driven learning models~\cite{naqvi2026ddosR}. Their results were instead used to identify the development of Urdu ASR, the effect of resource scarcity, and the remaining research problems. Fig.~\ref{fig:fig2} presents the review process, while Fig.~\ref{fig:fig3} presents the overall organization of the survey.

\section{AUTOMATIC SPEECH RECOGNITION SYSTEM FOR URDU}\label{sec:urdu-asr}

ASR approaches can generally be divided into monolingual and multilingual systems. A monolingual system is trained to recognize one language, whereas a multilingual system supports several languages within a shared model. This survey focuses on Urdu and examines both Urdu-specific systems and multilingual systems that report Urdu results. The reviewed monolingual methods are grouped into statistical, DNN-based, and end-to-end approaches. Multilingual and cross-lingual methods are discussed separately because they transfer information from other languages to Urdu. The literature shows that early Urdu ASR systems mainly relied on statistical methods, while recent studies have increasingly used deep learning, self-supervised learning, and multilingual pre-training.


Most early Urdu ASR studies used isolated words, spoken digits, or small-vocabulary read-speech datasets. Some datasets contain speech from a single speaker, while others include a limited set of sentences recorded by several speakers  \cite{ardila,amjad2020data, asghar, ashraf, baker}. For example, Qasim et al. \cite{qasim2016district} used the names of 139 districts of Pakistan, while Raza et al. \cite{raza2009phonetically} developed a corpus containing 250 isolated Urdu words. More recent studies have introduced larger continuous, conversational, and broadcast-speech datasets. However, many of these datasets are not publicly available. Table~\ref{tab:datasets} summarizes the Urdu speech datasets identified in the reviewed literature \cite{ali2015digits, baker,zia2019lstm}. The following subsections describe their main characteristics and their use in Urdu ASR research.


\begin{table}[htbp]
\centering
\caption{Urdu speech recognition systems and datasets}
\label{tab:datasets}
\scriptsize
\setlength{\tabcolsep}{3pt}
\renewcommand{\arraystretch}{1.5}
\begin{tabular}{@{}p{0.5cm} p{3.75cm} p{1.8cm} p{1.3cm} p{2.3cm} p{1.2cm} p{0.6cm}@{}}
\toprule
\textbf{S. No} & \textbf{Dataset} & \textbf{Size (utts./hrs)} & \textbf{Vocab} & \textbf{Speech Type} & \textbf{Public} & \textbf{Ref.} \\
\midrule
1 & Urdu Spoken Words & 5200 utts. & 52 & Isolated & No & \cite{ashraf} \\
2 & AUDD & 25,518 utts. & - & isolated & No & \cite{chandio} \\
3 & Isolated Urdu Corpus & 250 Words & 2500 & Isolated & Yes & \cite{ali2016corpus} \\
4 & PRUS Corpus & 70 Minutes & 42,289 & Read Speech & Yes & \cite{raza2010spontaneous} \\
5 & District Names DS & 9.5 hours & 41293 & Spontaneous, isolated & Yes & \cite{qasim2016district} \\
6 & Urdu Mixed Datasets & 126.8 hours & - & Read Speech & Partially & \cite{naeem} \\
7 & Common Voice-19.0 & 301 hours & - & Spontaneous & Yes & \cite{ardila} \\
8 & Large Vocab Urdu speech LVCSR & 300 hours & 199K & Read Speech & No & \cite{farooq2019dnn} \\
9 & RUMI dataset & 1.81 hours & 6693 & Interview & Yes & \cite{raza2010spontaneous} \\
10 & Spontaneous Speech Corpus & 45 hours & 14000 & Spontaneous, read speech & No & \cite{sarfraz} \\
15 & Telephone Speech & 1207 hours & 20K & Conversational & No & \cite{raza2018telephonic} \\
16 & Apptek Broadcast Data & 99 hours & 800K & Broadcast & No & \cite{shaik} \\
17 & Shrutilipi Dataset & 193 hours & 0.17 Million & Broadcast, News & Yes & \cite{bhogale2023mining} \\
18 & IndicSUPERB (Kathbath-Dataset) & 86.7 hours clean; 77.0 hours noisy & 44K & Read speech & Yes & \cite{javed} \\
19 & LDC for IL Urdu Data & 99.2 Hours & 57M & Read Speech & No & \cite{rao} \\
20 & Multi Genre Urdu Data & 98 hours & 200K & Broadcast & No & \cite{khan} \\
21 & Vistaar Benchmark & 320 hours & - & Combination & Partially & \cite{bhogale2023vistaar} \\
22 & CMU Wilderness & 19 hours & 2000 & Read Speech & Partially & \cite{black} \\
23 & FLEURS & 8.5 hours & - & Read Speech & Yes & \cite{conneau} \\
\botrule
\end{tabular}
\end{table}


\subsubsection{Audio Urdu Digits and Spoken Words Dataset}\label{subsubsec:audios}

\begin{figure}[htbp]
\centering
\includegraphics[width=0.65\textwidth]{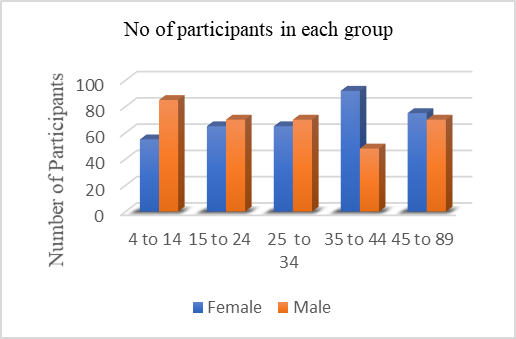}
\caption{Gender and Age of Participants}\label{fig:fig4}
\end{figure}

Research article Chandio et al., \cite{chandio} presents a novel work of Urdu Audio Digits (AUDD) datasets with multiple age groups and gender. The dataset contains 25,518 collected from 740 participants. The dataset has multiple different diversity classes that makes it more useful in order to train them for speech recognition. Fig.~\ref{fig:fig4} represents the AUDD datasets diversity on basis of age and gender. Similar distribution can be found across different datasets utilized for ASR research by different researchers. This dataset can be utilized to perform spoken Urdu digits recognition as the authors have performed using different baseline algorithms as well as proposed CNN based classification with comparative performance. Bit similar though in smaller size dataset developed and presented by \cite{ashraf}. Authors of the literature has mentioned a dataset of 52 Urdu spoken words by 10 speakers making a total of 5200 utterances in total, as a baseline of Speech recognition. Isolated Urdu words datasets focusing district names of Pakistan is introduced in literature \cite{qasim2016travel}. This dataset consists of 139 nouns spoken by 300 speakers, male and female, making a total of 41293 utterances and 9.5-hour duration. Furthermore, the dataset has the variety of speakers classified under Punjabi, Pashto, Sindhi, Balochi, Seraiki and native Urdu speakers.

\subsection{Urdu ASR Datasets}\label{subsec:datasets}

Another attempt of developing Urdu isolated words speech dataset is presented by \cite{ali2015digits}. This dataset consists of 250 isolated Urdu words each of which has been spoken by ten speakers, discriminated by gender, native and non-native speakers as well as by age groups.


\subsubsection{Rumi: Read and Spontaneous Speech Corpora}\label{subsubsec:rumi}

Initial work for spontaneous Urdu speech dataset and recognition system has been done and presented by \cite{raza2009phonetically} and extended into literature \cite{raza2010spontaneous}. This small Urdu transcribed speech dataset consists of 70 minutes, transcribed Urdu read speech from 708 greedily made sentences, is publicly available for research use only. The authors are of the view that this dataset, which consists of 10101 words and 5656 unique words containing 60 unique phones and 42289 phone occurrences represents all the phones and tri-phone combinations used in Urdu. A similar dataset of spontaneous Urdu speech presented by \cite{sarfraz}, consists of 45 hours spontaneous and read speech from 82 speakers exhibiting variety of speakers as per gender and age groups, but not available publicly.


\subsubsection{Large Vocabulary Urdu Speech}\label{subsubsec:large-vocab}

Research literature by Farooq et. al., \cite{farooq2019dnn} presents Urdu speech data of 300 hours covering a vocabulary size of 199K from 1671 speakers. This is a huge work compare to many existing Urdu datasets till now. The researchers though haven't yet contributed the dataset for research community. So it is a private dataset. Although an ASR system, trained with the said datasets using traditional approaches like HMM-GMM and TDNN-BLSTM networks, is available publicly as a web service.

An approach adopted in literature \cite{raza2018telephonic}, targeting telephone community of 11,017 users, gathered a large collection of 1207 hours of speech. This dataset comprises not only Urdu, but ranges through nine other regional languages. Speech dataset collected with this approach is not fully processed and adopted in ASR research, rather only a small proportion 9.5 hours tested using the baseline GMM and SGMM models and SRILM toolkit. Literature \cite{naeem} reported large vocabulary transcribed combination of different speech datasets already presented in literature like the CSALT dataset and PRUS corpus mentioned before making a 100 hour in total. Another dataset with large dataset is presented by \cite{shaik}. This Urdu speech dataset is unique in nature as it is collected from broadcast sources. The researchers conducted experiments using 99-hour speech and vocabulary size of 79K for which 266M words have been collected from multiple sources.


\subsubsection{Common voice}\label{subsubsec:common-voice}

Mozilla\textquotesingle s Common Voice project \cite{ardila}  is an open-source initiative to collect speech data from volunteers reported for language identification in literature by \cite{singh}. The project aims to make speech data more accessible and affordable for developers of speech-enabled applications. To date, Common Voice has gathered over 2,400 hours of speech data in over 15 languages. Of this data, over 1,900 hours have been validated. This means that the dataset has been checked to ensure that it is of high quality and can be used to train speech recognition models. The data collected by Common Voice is available to developers for free. This makes it a valuable resource for anyone who wants to develop speech-enabled applications. The Common Voice project is an important step towards making speech recognition technology more accessible and affordable. By making speech data more widely available, the project is helping to make it possible for developers to create new and innovative speech-enabled applications. So far the Common Voice Corpus 19.0 includes 301 recorded hours out of which 46 hours Urdu speech data is validated.


\subsubsection{Vistaar and Kathbath Datasets}\label{subsubsec:vistaar}

Shrutilipi dataset containing 6,400 hours labelled audio along with 4.95M huge sentences across 12 Indian languages, including Urdu presented in literature \cite{bhogale2023mining}. The publicly available Urdu speech dataset under this repository is 0.31K hours and 0.17M sentences. Another big attempt to collect and present multilingual dataset benchmark of major Indic languages including Urdu is presented in a recently published literature by \cite{bhogale2023vistaar}. Out of 10,736-hour speech dataset proportion of Urdu is 320-hours long. A huge amount of Indic Languages labelled Speech Dataset comprising 1684 hours, covering 12 major Indian languages from 1218 contributors is also presented in literature by \cite{javed}. This dataset includes two types of Urdu speech datasets: 86.7 hours clean and 77.0 hours noisy from 36 males and 31 female speakers respectively.


\subsubsection{FLEURS and Whisper Multilingual Speech Dataset}\label{subsubsec:fleurs}

With the claim of human-level accuracy, an open-source neural net is introduced by OpenAI recently \cite{radford}. This is a massive dataset consisting of 680,000-hour multilingual and multitask supervision from 96 different languages including Urdu. Urdu has overall 104-hour speech data for multilingual speech recognition in Whisper, though the dataset is not novel rather reuse of other datasets like FLEURS \cite{conneau} and Common Voice \cite{ardila} as per the ASR results mentioned in literature. FLEURS, a multi-lingual dataset of 102 languages, included 1.4 hours long Urdu read speech, justifiably treated Urdu as an unseen language as its been excluded in the pre-training phase. Yet the article \cite{conneau} describes that each language included in FLEURS has 12 hours of speech data. Based on the pre-trained Whisper model, multiple languages have been successfully fine-tuned achieving good ASR accuracies.


\subsubsection{Other Non-Commercial and Commercial Licensed Urdu Speech Datasets}\label{subsubsec:other-data}

Other than the datasets reported from literature above, there are several collections of huge amount Urdu speech datasets build by organizations and private groups which are available under commercial license. To be mentioned 99.2 hours of Urdu read speech for automatic speech recognition is developed and presented by \cite{rao}.

Apptek dataset is prepared specifically for the ASR system of 60+ Indian under resource languages. The dataset for Urdu language from this diverse collection has been used for ASR evaluation of Urdu and some European languages \cite{shaik}. It altogether contains 99 hours of broadcast Urdu news with 800K words and 29K segments (privately available at \href{http://www.appteck.com}{www.appteck.com}).

Multi-genre Urdu broadcast speech data is developed and introduced to the ASR research by \cite{khan}. This broadcast speech dataset, gathered from radio, YouTube, and TV sources, comprises a total of 98 hours as per the mentioned literature. The speakers are 453 in total out of which 333 are male and 120 are female speakers. Other variations in this dataset are categorized as politics, health, entertainment, current affairs, etc.

Urdu speech dataset is also included into the CMU Wilderness Multilingual speech dataset brought into publication by Alan et. al and presented in multiple scientific publications \cite{black,chen2023focal,chen2022meta}. This dataset consists of 700 different languages with the claim that each language has an average of 20 hours of speech data collected from bible.is. The Wilderness multiple speech dataset which is also named the Babel dataset though primarily utilized only for speech synthesis instead of automatic speech recognition; but multiple researchers have adopted it for various ASR tasks in publications by Chen et. al., \cite{chen2022meta, chen2023focal, thomas}.


\subsection{Tools and Software Used for Urdu ASR}\label{subsec:tools}

Early Urdu speech-recognition studies examined the recognition of spoken digits using MATLAB and the fast Fourier transform (FFT) \cite{hasnain}. Later statistical systems used toolkits such as Kaldi, Sphinx4, SRILM, and KenLM. Naeem et al. \cite{naeem}, for example, used Kaldi and KenLM for acoustic and language modeling. Ashraf et al. \cite{ashraf} used Sphinx4 with a word-list grammar for speaker-independent Urdu speech recognition. Sphinx was also used for spontaneous speech, district-name recognition, and small-vocabulary tasks \cite{raza2010spontaneous, shaukat, qasim2016district}. Kaldi and SRILM were commonly used in SGMM-, HMM-, and DNN-based Urdu ASR studies \cite{raza2018telephonic, farooq2019dnn, farooq2020code, khan}.

Table~\ref{tab:tools} summarizes the main tools, model types, and reported results. More recent studies have used self-supervised models such as wav2vec 2.0 and XLSR, which support multilingual pre-training and adaptation to low-resource languages \cite{mohiuddin}.

Monolingual ASR systems are designed to recognize speech in one language. Early systems commonly used hidden Markov models (HMMs) to represent the temporal sequence of speech units. HMM-based systems can perform well on small and carefully prepared datasets, but they are sensitive to noise, speaker variation, and pronunciation differences. Other traditional approaches used for Urdu include Gaussian mixture models (GMMs), support vector machines, random forests, linear discriminant analysis, and subspace Gaussian mixture models \cite{ali2016corpus, naeem, chandio}. Another approach is to use deep neural networks (DNNs) initially reported by researchers like \cite{mohamed}. DNNs are more powerful than GMMs and can learn to recognize speech even in the presence of noise and variations in pronunciation. However, DNNs are more complex to train and require larger datasets.

In recent years, there has been a trend towards using end-to-end ASR systems \cite{graves, wang}. Cited research literatures describe that end-to-end systems do not use a separate acoustic model and language model. Instead, they use a single neural network to map from audio to text. End-to-end systems have the potential to achieve higher performance than traditional ASR systems, but they are still under development. According to the literature \cite{zhang2017exploitation}, the choice of monolingual ASR approach depends on several factors, including the availability of data, the desired performance level, and the budget. For example, HMM-based systems are a good choice for low-resource languages or applications where accuracy is not critical. DNN-based systems are a good choice for high-resource languages or applications where accuracy is important whereas the End-to-end techniques are preferred where the highest possible performance is desired given resource-rich languages and computational resources.

Traditional techniques have been the most adopted ones across Urdu ASR, which might be because of the relevancy and adaptability of the limited-scale datasets. Although the accuracy of the experiments cannot be said to be much outstanding, but they are considered to be the basis and building blocks of later techniques with improved results. DNNs got so much attention in leveraging high accuracy for other languages with big datasets. E2E techniques also require a high volume of data to generate good accuracy transcriptions, yet in the case of Urdu ASR the datasets available are still far from the required, so we could rarely find a literary work to mention. Multilingual ASR assisted many low-resource languages including Urdu, with the advent of transfer learning and its variants. The following sub-sections summarize various ASR approaches, mainly utilized for Urdu language as a monolingual ASR, followed by the multilingual and transfer learning adaptation.


\subsubsection{Traditional HMM-GMM-based Urdu ASR approaches}\label{subsubsec:hmm-gmm}

As Urdu has been a resource-scarce language, the baseline automatic speech recognition so far is mainly with the traditional HMM-GMM-based approaches. Speech recognition framework CMU Sphinx4 mainly based on HMM is adopted by the researchers \cite{ashraf} over spoken Urdu isolated words dataset. Word List Grammar with 52 isolated words are used to develop the speech corpus from 10 different speakers. Similarly, Discrete Fourier Transforms DFT for Urdu ASR is described in the literature by \cite{beg, hasnain}. These literary works stand to be the initial attempts in Urdu ASR, though focus only the recognition of Urdu numbers 0 to 9 using MATLAB programming rather than any specified ASR toolkit.

As the literature presents \cite{naeem, shaukat}, the core objective of Automatic Speech Recognition is proposing a model of best possible distribution of word sequence. Sequential ASR system is based on estimating maximum a posteriori probability which need transforming the sequence of the acoustic characteristics into a word sequence; represented by $X=\{x_t\in\mathbb{R}^D\mid t=1,\ldots,T\}$ and $W=\{w_n\in V\mid n=1,\ldots,N\}$ respectively. When $x_t\in\mathbb{R}^D$ and $T$ represents the length of vectors, $N$ shows the length of the word sequence, whereas $V$ is vocabulary, the maximum probability of word sequence is estimated for all vocabularies $W$ as:

\begin{equation}
W^{*}=\arg\max P
\label{eq:map-word}
\end{equation}

The objective of this sequential ASR system involves a pipeline of several processes under formal statistical approach: Feature extraction from speech input, acoustic modeling, language modeling, and decoding the word sequence. Different mechanisms are adopted by researchers in order to calculate the said models, like HMM-GMM, HMM-DNN for acoustic modeling, and RNN, and LSTM for language modeling for varying languages and datasets depending on the language and dataset.

Literature \cite{shaukat} presents HMM based ASR approach for medium vocabulary Urdu dataset consisting 250 words. This initial level work is done using CMU Sphinx again with reportedly better results. Discrete Fourier Transforms DFT for feature extraction and Feed Forward Artificial Neural Networks FANN for classification combining with Discrete Wavelet Transforms DWT reported by  \cite{rehmam}. According to the literature, this approach proves better performance as compared to baseline ASR experiments. ~Subspace Gaussian Mixture Model SGMM is adopted to examine the novel collection of spontaneous Urdu speech corpora in \cite{raza2018telephonic} integrated with SRILM toolkit for the Kneser-Ney discounting-based trigram language model. GMM and SVM with Random Forest approaches have been adopted for accent identification across Urdu speech by authors in the literature  \cite{qasim2016travel}. Researchers of the literature \cite{shaik} also relied upon adopting the traditional HMM approach for Urdu LVCSR in their evaluation system presented for multiple languages.

HMM mechanism is utilized both in acoustic and language models, though it is advantageous modeling with the acoustic modeling part. HMM-based model utilizes Bayesian theorem to describe HMM state sequence $S=\{s_t\in\{1,\ldots,J\}\mid t=1,\ldots,T\}$, for observation $X$ and latent state set $\{1,\ldots,J\}$. Hence the acoustic, language, and pronunciation models tend to give the probability of observing $X$ from hidden sequence $S$ as:

\begin{equation}
\underset{L\in\gamma^{*}}{\arg\max}\,p(L\mid X)
\approx
\underset{L\in\gamma^{*}}{\arg\max}
\sum_{S}p(X\mid S)p(S\mid L)p(L)
\label{eq:hmm}
\end{equation}

Convolutional Neural Networks CNN and Support Vector Machine SVM were used for the recognition of Urdu audio digits' recognition in literature \cite{chandio}. Different from the discussed techniques has been presented in the literature by  \cite{zia2019lstm}. Deep bi-directional LSTM architecture is used in combination with the features of contextual information generating complete context of the sequence. The authors argue that a higher accuracy is achieved this way as a comparison is made with gated recurrent unit GRU-based architectures. The Subspace Gaussian Mixture Model SGMM technique adopted for spontaneous Urdu speech data collected from telephonic forum \cite{raza2018telephonic} and achieved remarkable WER. A similar approach applied for continuous Urdu speech recognition in literature \cite{naeem}. Results show that the SGMM approach in this research is better than the conventional GMM-HMM baseline.

For multi-genre Urdu broadcast speech Time delay neural network TDNN is applied \cite{khan}. This application of TDNN with RNN language model is for the first time in Urdu ASR, also for a large amount of broadcast speech data with multi-genre, so the the study reported improved results. Table~\ref{tab:statistical} outlines the statistical techniques and results of ASR along with utilized datasets by different researchers.


\begin{table}[htbp]
\centering
\caption{Statistical Techniques and Results Summary}
\label{tab:statistical}
\scriptsize
\setlength{\tabcolsep}{3pt}
\renewcommand{\arraystretch}{1.25}
\begin{tabular}{@{}p{0.7cm} p{3cm} p{2cm} p{0.7cm}@{}}
\toprule
\textbf{S. No} & \textbf{Technique(s)} & \textbf{\%WER/ Accuracy} & \textbf{Ref.} \\
\midrule
1 & SGMMI & 16.7 & \cite{naeem} \\
2 & SVM, CNN & 0.97 & \cite{chandio} \\
3 & GMM, CMLLR & 32.6 & \cite{shaik} \\
4 & SVM, RF, LDA & 73\% & \cite{ali2015digits} \\
5 & HMM & 74\% & \cite{shaukat} \\
6 & SVM, GMM & 92.4\% & \cite{qasim2016travel} \\
7 & SGMM, MMI & 24.2 & \cite{raza2018telephonic} \\
8 & HMM & 10.66 & \cite{ashraf} \\
\botrule
\end{tabular}
\end{table}


\subsubsection{DNN-based Urdu ASR approaches}\label{subsubsec:dnn}

Semi-supervised deep learning techniques have been used over the isolated words dataset, presented earlier by researcher \cite{ashraf}, and in literature \cite{humayun}. The model generalization improvement technique adopted has successfully achieved better performance as compared to the traditional sphinx-based recognition by the former. Literature by \cite{farooq2019dnn} implemented Urdu automatic speech recognition using a large vocabulary with deep neural networks. Along with baseline GMM, the researchers implemented Time Delay Neural Networks TDNN, Long Short-term Memory LSTM and Bidirectional LSTM networks proven to be giving high accuracy in terms of word error rate WER when implemented a with recurrent neural network RNN RNN-based language model. DNN-HMM-based approaches were also adopted by \cite{farooq2020code} for a combination of Urdu broadcast speech and spontaneous speech datasets without prior GMM-HMM training and alignments. This Deep neural net application successfully reduced the word error rate compare to other similar applications implementing Urdu-English conversational code-switched speech as well. Figure 7 shows the summary of deep neural network DNN-based approaches adopted so far for Urdu speech recognition research.

As The GMM-based approach is successful enough for direct acoustic modeling from input sequences, simulating every individual state distribution. The ability to give conditional probability directly makes the Deep neural networks DNNs more efficient than GMMs. Probability model $P(x_t\mid s_t)$ is replaced with a classification problem $P(s_t\mid x_t)$ in DNN by utilizing pseudo-pseudo-probability mechanism as a joint probability approximation. Mechanism of DNN performing frame-by-frame training from frame-by-frame alignment with input $x_t$ and target state $s_t$, prior portability of state $P(s_t)$ in the denominator $p(s_t)$, is mathematically described under:

\begin{equation}
\prod_{t=1}^{T}P(x_t\mid s_t)
=
\prod_{t=1}^{R}\frac{P(x_t\mid s_t)}{p(s_t)}
\label{eq:dnn}
\end{equation}

Semi-Supervised Deep Learning SSL approach featured with unsupervised dimensionality reduction technique i.e. Locally Linear Embedding LLE, also been adopted for Urdu speech recognition \cite{humayun}. Despite an insufficient labeled dataset, the proposed semi-supervised deep learning approach proved to be having comparable results with the HMM-based benchmark on a similar amount of data. Another technique in a recently published scholarly work \cite{javed} shows that self-supervised learning with Wav2vec2 approach also stands fascinating in terms of the monolingual model while adopting larger Urdu language dataset.


\subsubsection{End-to-End Architecture (E2E) for ASR}\label{subsubsec:e2e}

The term end-to-end in the context of Automatic Speech Recognition (ASR) has several interpretations. It involves joint modeling, where all ASR components are treated as a unified computational graph, unlike traditional distinctions of acoustic and language models. Meaning, the end-to-end models simplify an ASR by directly mapping acoustic input to transcriptions without the need for explicit alignments or discrete acoustic and language models. A single-pass search integrates all ASR components before making decisions, aligning with Bayes\textquotesingle{} decision rule to integrate possible resources and outputs the result. The E2E paradigm estimates model parameters minimizing the WER. The training uses a single data type, typically speech audio data, but must address challenges of un-transcribed audio and text-only data. End to End is also described as its ability to train from scratch requires training without external knowledge, particularly important in scenarios like self-supervised learning \cite{luo}.

\begin{figure}[htbp]
\centering
\includegraphics[width=0.95\textwidth]{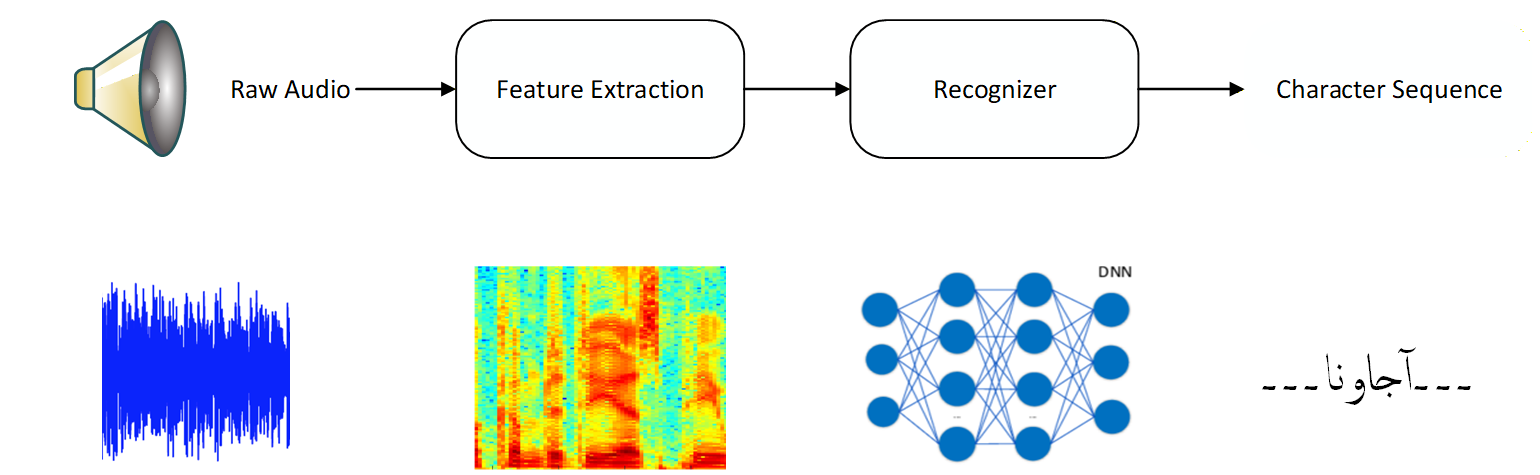}
\caption{General architecture of end-to-end ASR system}\label{fig:fig5}
\end{figure}

Furthermore, E2E ASR avoids reliance on secondary knowledge sources such as pronunciation lexica and phonetic clustering, which can introduce errors and costs. Vocabulary modeling without lexica may limit vocabularies, with whole-word models demanding extensive training data. E2E ASR pertains to whether task-specific constraints are learned from data or directly implemented like attention-based models attempt learning from data whereas classical HMM architecture follows a different flow. In short, E2E ASR represents integrated modeling with joint training and recognition, reducing complexity and improving genericity \cite{chan, adjoudani}. Fig.~\ref{fig:fig5} demonstrates the general architecture of an end-to-end automatic speech recognition system.

Research literature \cite{graves, luo, wang, watanabe2018espnet} presents the calculation of $P(W\mid X)$ with end to end model, acoustic features input $X=(x_1,\ldots,x_t)$ with sequence of targets as $y=(y_1,\ldots,y_t)$ and word sequence as $W=w_m=(w_1,\ldots,w_m)$, likelihood $P(\cdot\mid x_1),\ldots,P(\cdot\mid x_t)$ is calculated with an artificial neural network ANN using input labels given as word representations. Large datasets form the basis of E2E ASR models, hence it\textquotesingle s unlikely to get high accuracy with E2E for low-resource languages.


\subsection{Multilingual ASR with Urdu Language}\label{subsec:multilingual}

Despite being an under-resource language, researchers have adopted Urdu language for automatic speech recognition task together with different other languages. Literature by \cite{javed} developed and presented their work focusing multiple languages spoken across India included Urdu. Major task in this research work is to develop speech datasets, though they have presented their experimental work to test the dataset using self-supervised learning SSL approach. ASR is not the solely task of this literature rather speaker verification, language understanding, speaker identification, and keyword spotting is also done for 12 Indic Languages including Urdu. In this task, the acoustic model is built separately instead of the joint multilingual model.

A large-scale dataset for Indic languages was presented by \cite{bhogale2023mining} for 12 different languages spoken in India including Urdu. Wav2Vec and Conformer models have been adopted along with KenLM for several Indic languages including Urdu language, and the results shown are comparatively better than another benchmark. In order to improve the dataset and accuracy of ASR for several Indic languages another attempt was recently reported in literature by \cite{bhogale2023vistaar}. This piece of scientific work presents Vistaar, a set of 59 benchmarks across various languages and domain combinations including Urdu language. This benchmark is a combination of multiple high-volume datasets and benchmarks across languages making 10,736 hours in total and specifically 320 hours in Urdu alone. The evaluation results show that there is a big reduction in terms of WER.

Another major attempt with massive datasets, claiming to consist of 680,000 hours mainly of English speech and 117,000 hours across multilingual ASR data is reported in research by Radford et al. \cite{radford}, a very well-known OpenAI's Whisper ASR system. Basically, Whisper is a transformer architecture, well suited for sequential tasks like automatic speech recognition and, pre-trained model for automatic speech recognition transcribing speech in a variety of languages, including Urdu. Urdu language is part of Whisper with hours from multiple sources. Whisper initially converts the speech audio signals to a sequence of Mel spectrograms, representing the frequency content of audio signals. The transformer architecture then learns and maps the sequence of spectrograms to a sequence of text tokens, which are converted to string text as transcribed speech. This state-of-the-art results on a variety of ASR benchmarks including the Urdu ASR benchmark with 17.5\% WER with comparatively high volume dataset. A summary of multilingual speech recognition systems which include the Urdu language are presented in Table~\ref{tab:multilingual}.

\begin{table}[htbp]
\centering
\caption{Multilingual ASR Systems with the Urdu Language}
\label{tab:multilingual}
\scriptsize
\setlength{\tabcolsep}{3pt}
\renewcommand{\arraystretch}{1.5}
\begin{tabular}{@{}p{0.7cm} p{4.5cm} p{2cm} c c p{0.7cm}@{}}
\toprule
\textbf{S. No} & \textbf{A.M Technique(s)} & \textbf{Language Model LM} & \shortstack{\textbf{WER (\%)}\\\textbf{Urdu}} & \shortstack{\textbf{Avg WER (\%)}\\\textbf{(Multi Lingual)}} & \textbf{Ref.} \\
\midrule
1 & SSL, IndicWav2Vec2, XLS-R & KenLM & 12.6 & 12.4 & \cite{javed} \\
\addlinespace[2pt]
2 & Wav2Vec, Conformer, CTC, OpenSLR & KenLM & - & 15.3 & \cite{bhogale2023mining} \\
\addlinespace[2pt]
\multirow{4}{*}{3} & IndicWav2Vec & \multirow{4}{*}{-} & 25.1 & 21.0 & \multirow{4}{*}{\cite{bhogale2023vistaar}} \\
& IndicWhisper, etc. & & 19.4 & 13.6 & \\
& Google SST & & 23.3 & 23.9 & \\
& Azure SST & & - & 20.0 & \\
\addlinespace[2pt]
\multirow{2}{*}{4} & Whisper Model: Seq-to-Seq Transformer on FLEURS & \multirow{2}{1.8cm}{Audio conditional LM} & 22.6 & - & \multirow{2}{*}{\cite{radford}} \\
& Whisper Model: Seq-to-Seq Transformer on Common Voice9 & & 24.2 & - & \\
\addlinespace[2pt]
\multirow{2}{*}{5} & FLEURS: Wav2Vec-Bert & \multirow{2}{*}{-} & \shortstack{82.9\\(\% CER$\downarrow$)} & \shortstack{14.1\\(\% CER$\downarrow$)} & \multirow{2}{*}{\cite{conneau}} \\
& FLEURS: mSLAM & & \shortstack{83.1\\(\% CER$\downarrow$)} & \shortstack{14.6\\(\% CER$\downarrow$)} & \\
\botrule
\end{tabular}
\end{table}


\subsection{Fine Tuning XLSR for Urdu}\label{subsec:xlsr}

\begin{figure}[htbp]
\centering
\includegraphics[width=0.98\textwidth]{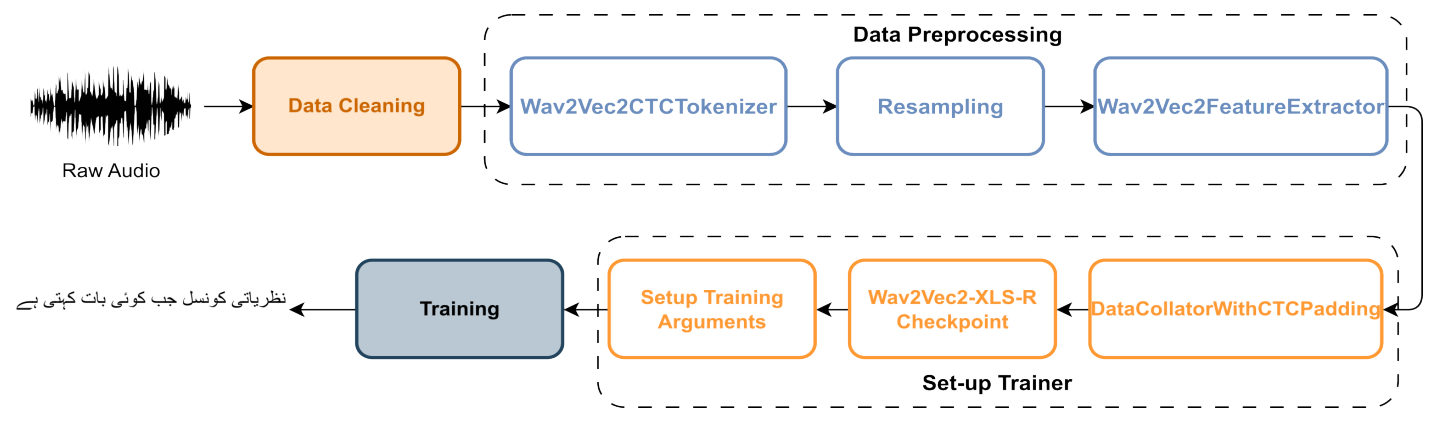}
\caption{Wav2Vec2.0 paradigm adopted for Urdu ASR \cite{mohiuddin}}\label{fig:fig6}
\end{figure}

With the advent of self-supervised learning algorithms, large-scale pre-training, fine-tuning, and diverse applications of Transformers, LLMs are contributing to and benefitting ASR research \cite{fathullah}. From recent studies, it is also evident that ASR of languages with big datasets achieves human-level parity with sophisticated algorithms like transformers in LaMDA \cite{thoppilan}, OPT \cite{zhang2022opt}, and Whisper \cite{radford}. A recent conference paper, which is the sole study focusing particularly on Urdu Language, adopted pre-trained XLSR model to fine-tune Urdu speech from Common Voice \cite{mohiuddin}. The accuracy achieved this way, as presented in Table~\ref{tab:tools} and the illustration of methodology in Fig.~\ref{fig:fig6}, stands comparable and better in some cases of low resource language ASR, yet the way it is attempted is novel for the case of the Urdu language.

\section{COMPARISON OF METHODS ADOPTED IN URDU ASR}\label{sec:comparison}

Statistical methods, particularly HMM-based systems, have been widely used in Urdu ASR because they can be trained on relatively small datasets. As larger datasets became available, researchers introduced DNN, TDNN, LSTM, and BLSTM models to learn more complex speech patterns. These models generally reduced WER compared with conventional statistical baselines, although their results remain dependent on the dataset, speech type, and evaluation setting.

\begin{figure}[htbp]
\centering
\includegraphics[width=0.95\textwidth]{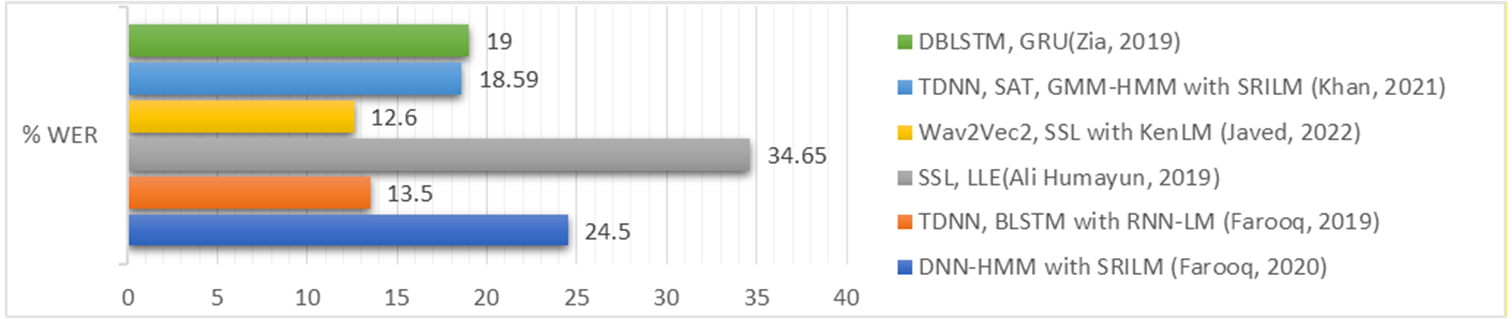}
\caption{DNN based approaches for Urdu ASR\cite{zia2019lstm,khan,javed,humayun,farooq2020code,farooq2019dnn}}\label{fig:fig7}
\end{figure}

Table~\ref{tab:datasets} summarizes the reviewed Urdu speech datasets. Table~\ref{tab:tools} lists the main software tools, model types, and reported performance measures, while Table~\ref{tab:statistical} focuses on statistical approaches. Fig.~\ref{fig:fig7} summarizes the results reported for DNN-based systems. Because these studies use different datasets and experimental settings, their reported results should not be treated as direct model-to-model comparisons.

Tables~\ref{tab:multilingual} and \ref{tab:multilingual-comparison} summarize multilingual systems that include Urdu. WER is the most frequently reported measure, although some studies use CER. Among the reviewed multilingual studies, IndicWav2Vec with self-supervised learning and KenLM reported an average WER of 12.4\% and an Urdu-specific WER of 12.6\% \cite{javed}. These results indicate the potential of multilingual pre-training for Urdu ASR.


\begin{table}[htbp]
\centering
\caption{Comparison of multilingual ASR system with Urdu language}
\label{tab:multilingual-comparison}
\scriptsize
\setlength{\tabcolsep}{3pt}
\renewcommand{\arraystretch}{1.10}
\begin{tabular}{@{}p{0.7cm} p{2.0cm} p{2.7cm} p{2.9cm} p{2.5cm}@{}}
\toprule
\textbf{Ref.} & \textbf{Method} & \textbf{Objective} & \textbf{Accomplishments} & \textbf{Limitation} \\
\midrule

\cite{javed} &
IndicSUPERB\newline Wav2Vec2-based multimodal ASR &
\textbullet~Larger dataset, 1684 hours, for 12 Indic languages including Urdu.\newline
\textbullet~Benchmarks for speech tasks including ASR.\newline
\textbullet~Train and evaluate self-supervised models along with FBANK. &
\textbullet~Able to reduce average WER for average and specific language.\newline
\textbullet~Able to boost the accuracy with a Jointly fine-tuned multilingual model &
\textbullet~Limited to 12 Indic languages.\newline
\textbullet~Test results with unknown speakers are different from those with known speakers.\newline
\textbullet~Not robust in noisy environment \\

\cite{bhogale2023mining} &
Shrutilipi dataset\newline Wav2vec2-based multimodal ASR &
\textbullet~Proposed mining text and audio pairs of low-resource Indic languages, Shrutilipi 6400 hours, as an alternative to creating labeled datasets.\newline
\textbullet~Adopt CTC with the Needleman-Wunsch algorithm. &
\textbullet~Able to reduce WER by 5.8\% for 7 languages, can further with the Conformer model.\newline
\textbullet~High quality and diverse dataset effective for ASR down streaming.\newline
\textbullet~Able to Train models which are more robust to noisy input. &
\textbullet~Limited to 12 Indic languages\newline
\textbullet~Smallest proportion of Urdu.\newline
\textbullet~Unable to show language-specific accuracy \\

\cite{bhogale2023vistaar} &
Vistaar\newline IndicWhisper &
\textbullet~A set of 59 benchmarks collated from ASR systems named Vistaar.\newline
\textbullet~trained for 10.7K hours long 12 languages. &
\textbullet~Diversity in every language.\newline
\textbullet~Reduce an average of 4\% WER.\newline
\textbullet~High volume Urdu speech data 320 hours with 13.6\% WER. &
\textbullet~More datasets are possible to compete with high-resource languages and multi-lingual benchmarks. \\

\cite{radford} &
Whisper:\newline Wav2Vec and Seq-to-seq architecture &
\textbullet~Multilingual and multi-task 680,000 hours labeled datasets.\newline
\textbullet~Whisper Model: Robust speech processing with pre-trained Models based on Seq-to-seq transformer architecture. &
\textbullet~Capable of getting an Average 29.3\% WER on wav2vec without the language model, and 12.8\% WER on the Whisper model.\newline
\textbullet~Whisper model robust enough for long-form (few minutes to few hours) transcriptions. &
\textbullet~Low accuracy for low-resource languages.\newline
\textbullet~Highest proportion of English\newline
\textbullet~Lacks fine-tuning\newline
\textbullet~Strong decoder but no CTC or wav2vec together with a language model. \\

\cite{conneau} &
FLEURS\newline Few-shot learning, mSLAM, and wav2vec-Bert Models &
\textbullet~Consists of 102 languages with a 12-hour dataset per language including Urdu.\newline
\textbullet~Attempt to catalyze resource-scarce languages in speech tech technology. &
\textbullet~Simple, baselines for the ASR tasks on pre-trained mSLAM and wav2vec-Bert models.\newline
\textbullet~Considerable \%CER specifically in Urdu with \%CER reductions of 82.9 and 83.1 respectively. &
\textbullet~Limited datasets\newline
\textbullet~No Language model \\

\botrule
\end{tabular}
\end{table}


\section{DISCUSSION}\label{sec:discussion}

This study delves into the prevailing trends in automatic speech recognition (ASR) research within the context of the resource-scarce Urdu Language. Establishing a robust ASR model for Urdu is of paramount importance, necessitating the exploitation of spectral speech features and the integration of cutting-edge deep learning techniques, particularly deep neural network architectures trained on language-specific datasets. To the best of our knowledge, it offers a comprehensive overview of the diverse techniques employed in Urdu ASR research. The insights derived and summarized in this review are poised to captivate the attention of researchers and developers dedicated to advancing ASR for Urdu, thereby contributing into the progress of this domain without the necessity of delving into detailed and practically obsolete foundational studies. For example, the progression of ASR research with state-of-the-art high-accuracy algorithms often encounters incompatibility issues when interfacing with toolkits designed for small-scale datasets and fundamental machine learning algorithms. Furthermore, earlier stages of ASR systems require numerous utilities that were previously achievable only through manual processes. One notable challenge lies in creating language model requirements, which has traditionally been a labor-intensive and time-consuming task without the aid of automated tools.

Data is an evitable ingredient of contemporary speech research. So this study has presented almost all the developed datasets for Urdu ASR. Unlike researchers focusing on English and Mandarin, Urdu is not that focused on speech recognition. However, the presented datasets have been and are utilized for numerous applications that facilitate real-life scenarios. Application aspects of speech research are complete areas that can be described in a separate literature.

Recognition tools and software are numerous, like numerous algorithms with different accuracies and computational requirements. To the best of our knowledge, this study has listed out and presented the majority of algorithms and tools that researchers adopt. Presenting them within a single review will be advantageous for early-career researchers entering low-resource ASR, establishing a standardized baseline for future parameter-efficient adaptation strategies. With time, data development and advancements in tools and algorithms are inevitable. Several areas are there to broaden Urdu speech research like data development, adaptation of deep neural networks, high accuracy algorithms, techniques like transfer learning and adaptation, and developing more effective language models compatible with Persian script, which can boost the Urdu speech accuracy.

\begin{itemize}
\item To the best of our knowledge, this work provides the first comprehensive overview of diverse techniques, tools, and algorithms adopted in advancing Urdu ASR research.
\item It discusses the various Urdu speech datasets in terms of their volume, type, availability, and applicability.
\item It also offers a thorough comparison of ASR approaches, ranging from the early statistical methods used in Urdu ASR to contemporary Transformer models.
\item Monolingual and multilingual ASR studies are summarized, highlighting both their accomplishments and limitations.
\item This study also points out the lack of focus by the Urdu research community, particularly those associated with speech and language, noting that the development of Urdu ASR still appears to be lagging behind that of several other low-resource languages.
\item Addressing speech varieties like different dialects, age and gender-based variations, exploring local utilization of speech data and their adaptation in various applications and digital services are still beyond the scope, given the available data and contemporary research landscape.
\end{itemize}

The insights derived and summarized in this review are poised to captivate the attention of researchers and developers dedicated to advancing speech research and contribute into the progress of this domain without the necessity of delving into detailed and practically obsolete foundational studies. For example, the progression of ASR research with state-of-the-art high-accuracy algorithms often encounters incompatibility issues when interfacing with toolkits designed for small-scale datasets and fundamental machine learning algorithms. Furthermore, earlier stages of ASR systems require numerous utilities that were previously achievable only through manual processes. One notable challenge lies in the creation of language model requirements, which has traditionally been a labor-intensive and time-consuming task without the aid of automated tools.

The current landscape of Urdu Automatic Speech Recognition (ASR) research predominantly centers around constrained datasets, encompassing isolated words, numerical digits, sentences, and modest-sized read speeches. These datasets frequently feature solitary speakers, characterized by a limited pool of sentences or variations of akin phrases articulated by a multitude of speakers. Although commendable endeavors have been undertaken to introduce a semblance of diversity in these datasets, such as the Audio Urdu Digits and Spoken Words Dataset (AUDD) and datasets tailored to district names, there exists ample opportunity for the broadening and enrichment of Urdu speech datasets.

Notably, while substantial, larger-scale Urdu speech datasets have been established in later studies as presented in section 4.1, the majority of the referenced datasets remain under proprietary restrictions. The Common Voice initiative, an invaluable resource within the global speech data collection landscape, notably encompasses Urdu and is instrumental in promoting open access to speech data resources. Moreover, ongoing research endeavors are directed toward the discernment of emotional nuances within Urdu speech. Various datasets and initiatives, including Kathbath and IndicSUPERB, significantly contribute to the repository of Urdu speech data available for ASR research purposes. It is noteworthy that several commercially licensed Urdu speech datasets are accessible, such as those curated and archived by the Language Data Consortium (LDC), Speech Ocean, and SLE Lahore, albeit with associated financial considerations. In summary, this deduction underscores the compelling necessity for the cultivation of extensive and diverse Urdu speech datasets, thereby propelling the advancement of ASR technology in the Urdu language, aligning with the rigorous standards of Nature publications.

Large vocabulary Urdu ASR has witnessed substantial advancements through the adoption of deep neural networks (DNNs), including TDNN, LSTM, and Bidirectional LSTM networks, in conjunction with recurrent neural network (RNN)-based language models, resulting in markedly reduced word error rates (WER) compared to GMM-based approaches. Additionally, the application of DNN-HMM techniques to a combination of Urdu broadcast and spontaneous speech datasets, without prior GMM-HMM training, has further reduced word error rates, even in scenarios involving Urdu-English conversational code-switched speech.

Furthermore, the implementation of semi-supervised deep learning, augmented by unsupervised dimensionality reduction techniques such as Locally Linear Embedding (LLE), has showed comparable performance to HMM-based benchmarks, even when confronted with limited labeled data. In a recent development, self-supervised learning using the Wav2vec2 approach has showcased remarkable potential when applied to larger Urdu language datasets, particularly in the context of monolingual models. In summary, these observations underscore the substantial impact of recently implemented algorithms on Urdu speech recognition, underscoring their key role in enhancing accuracy and performance across a diverse array of datasets, thereby propelling the progress of ASR technology within the Urdu language domain.


\section{OPEN PROBLEMS AND CHALLENGES}\label{sec:open-problems}


\subsection{Limitations}\label{subsec:limitations}

This review summarizes the main datasets, methods, tools, and results reported in Urdu ASR research. However, it has several limitations. Some Urdu datasets and systems are privately owned, and complete technical information is not always publicly available. The reviewed studies also use different speech types, dataset sizes, and evaluation settings, which limits direct performance comparison. In addition, multilingual studies often report limited Urdu-specific information.

The review mainly focuses on ASR datasets, recognition models, and reported performance. It does not provide an experimental evaluation of the reviewed systems. It also does not cover all related tasks, such as speech synthesis, speaker recognition, and speech-emotion recognition, unless they provide data or findings relevant to Urdu ASR.


\subsection{Problems and Challenges}\label{subsec:problems-challenges}

Despite recent progress, several challenges continue to limit Urdu ASR. These challenges are summarized below.

\begin{itemize}
\item Limited speech data: Large, diverse, and publicly available Urdu speech datasets remain limited. Many existing datasets contain isolated words, read speech, or recordings from a small number of speakers.
\item Limited use of transfer learning: Few studies have examined the full potential of multilingual pre-training, transfer learning, and fine-tuning for Urdu. Further evaluation is needed across different datasets and speech conditions.
\item Dialect and pronunciation variation: Urdu is spoken with different regional accents and dialects. Models trained on one speaker group may not generalize well to other groups.
\item Code-switching: Urdu speakers frequently combine Urdu with English and regional languages. However, limited code-switched datasets are available for training and evaluating ASR systems.
\item Inconsistent evaluation: The reviewed studies use different datasets, tasks, and evaluation measures. This variation makes direct comparison difficult.
\item Model robustness: Future Urdu ASR systems should also be evaluated under noise, domain shift, and small input perturbations. Studies in other machine-learning applications have shown that changes to learned representations can reduce model reliability~\cite{naqvi2025adversarial}.
\item Limited application-oriented research: Urdu ASR has received limited attention in areas such as education, accessibility, media transcription, conversational systems, and the digital processing of Urdu literary content.
\end{itemize}

These limitations provide clear directions for future research. Progress will require open datasets, shared evaluation protocols, stronger multilingual adaptation, and greater coverage of natural and diverse Urdu speech.


\section{CONCLUSION}\label{sec:conclusion}

This paper reviewed the development of Urdu automatic speech recognition from statistical methods to DNN-based, end-to-end, self-supervised, and pre-trained multilingual models. The review covered Urdu speech datasets, feature extraction methods, recognition tools, model architectures, and evaluation measures. It showed that early Urdu ASR studies mainly used small and task-specific datasets, which supported statistical models but limited the use of data-intensive neural architectures.

Recent multilingual datasets and pre-trained models, including wav2vec 2.0, XLSR, IndicWav2Vec, and Whisper, provide useful opportunities for improving Urdu ASR. However, the field still faces limitations related to public data availability, dialect diversity, code-switching, speaker variation, and inconsistent evaluation. Future work should focus on developing large and diverse open datasets, adapting multilingual models to Urdu, and evaluating systems under common and clearly reported conditions. These steps can support the development of more accurate and practical Urdu ASR systems.

\section*{CONFLICT OF INTEREST}

The Authors declare that no competing financial or personal conflict could have appeared to influence the work reported in this paper.

\section*{AUTHOR CONTRIBUTION}

Mr. Muhammad Sharif conceptualized the review scope, performed the systematic literature search, and wrote the original draft. Dr. Sardar Shan Ali Naqvi contributed to the study screening, data extraction from selected articles, and the organization of review materials. Mr. Zeeshan Abbas contributed to the literature collection and reference curation. Prof. Ying Wu co-supervised the project.

\section*{Funding}
This research did not receive any specific grant from funding agencies in the public, commercial, or not-for-profit sectors.

\section*{DATA AVAILABILITY}

As the study conducted in this paper is a survey of numerous research studies, the very first survey of its kind in the particular Urdu ASR, so there is no such data that can be said it produced and availability
required.

\bibliography{sn-bibliography}

\end{document}